\documentclass[11pt]{article}

\usepackage[preprint]{acl}

\usepackage{times}
\usepackage{latexsym}
\usepackage[utf8]{inputenc}
\usepackage{geometry}
\usepackage{booktabs}
\usepackage{multirow}
\usepackage{amsmath}
\usepackage{amssymb}
\usepackage{diagbox}
\usepackage{threeparttable}
\usepackage{tabularx} 
\usepackage{fontawesome5}
\usepackage[most]{tcolorbox}
\usepackage{xcolor}
\usepackage{enumitem} 
\usepackage{colortbl} 
\PassOptionsToPackage{table}{xcolor}
\PassOptionsToPackage{dvipsnames}{xcolor}

\definecolor{boxheader}{RGB}{52, 76, 183} 
\definecolor{boxfill}{RGB}{244, 244, 244} 

\newtcolorbox{promptbox}[1][]{
  enhanced,
  breakable,                
  colback=boxfill,         
  colframe=boxheader,       
  coltitle=white,          
  title=\textbf{Prompt},    
  fonttitle=\large\bfseries,
  sharp corners,            
  boxrule=1pt,             
  left=5mm, right=5mm, top=5mm, bottom=5mm, 
  #1
}
\usepackage{float}

\usepackage[T1]{fontenc}

\usepackage[utf8]{inputenc}

\usepackage{microtype}

\usepackage{inconsolata}

\usepackage{graphicx}

\title{Behavior-Aware Item Modeling via Dynamic Procedural\\Solution Representations for Knowledge Tracing}

\author{
  Jun Seo\textsuperscript{1}\thanks{Equal contribution.} \quad
  Sangwon Ryu\textsuperscript{1}\footnotemark[1] \quad
  Heejin Do\textsuperscript{3}\thanks{Corresponding authors.}\quad
  Hyounghun Kim\textsuperscript{1,2}\quad
  Gary Geunbae Lee\textsuperscript{1,2}\footnotemark[2]
  \\
  \textsuperscript{1}GSAI, POSTECH \quad
  \textsuperscript{2}CSE, POSTECH \\
  \textsuperscript{3}ETH Zurich, ETH AI Center \\
  \texttt{\{sjin4861, ryusangwon, h.kim, gblee\}@postech.ac.kr} \\
  \texttt{heejin.do@ai.ethz.ch}
}
\begin{document}
\maketitle

\begin{abstract}

Knowledge Tracing (KT) aims to predict learners' future performance from past interactions. While recent KT approaches have improved via learning item representations aligned with Knowledge Components, they overlook the procedural dynamics of problem solving. We propose Behavior-Aware Item Modeling (BAIM), a framework that enriches item representations by integrating dynamic procedural solution information. BAIM leverages a reasoning language model to decompose each item's solution into four problem-solving stages (i.e., understand, plan, carry out, and look back), pedagogically grounded in Polya’s framework. Specifically, it derives stage-level representations from per-stage embedding trajectories, capturing latent signals beyond surface features. To reflect learner heterogeneity, BAIM adaptively routes these stage-wise representations, introducing a context-conditioned mechanism within a KT backbone, allowing different procedural stages to be emphasized for different learners. Experiments on XES3G5M and NIPS34 show that BAIM consistently outperforms strong pretraining-based baselines, achieving particularly large gains under repeated learner interactions. Code and data are available in: \href{https://github.com/sjin4861/Behavior-Aware-Item-Modeling}{\faGithub\ sjin4861/BAIM}.

\end{abstract}
\section{Introduction}
\begin{figure}[t]
    \centering
    \includegraphics[width=0.85\columnwidth]{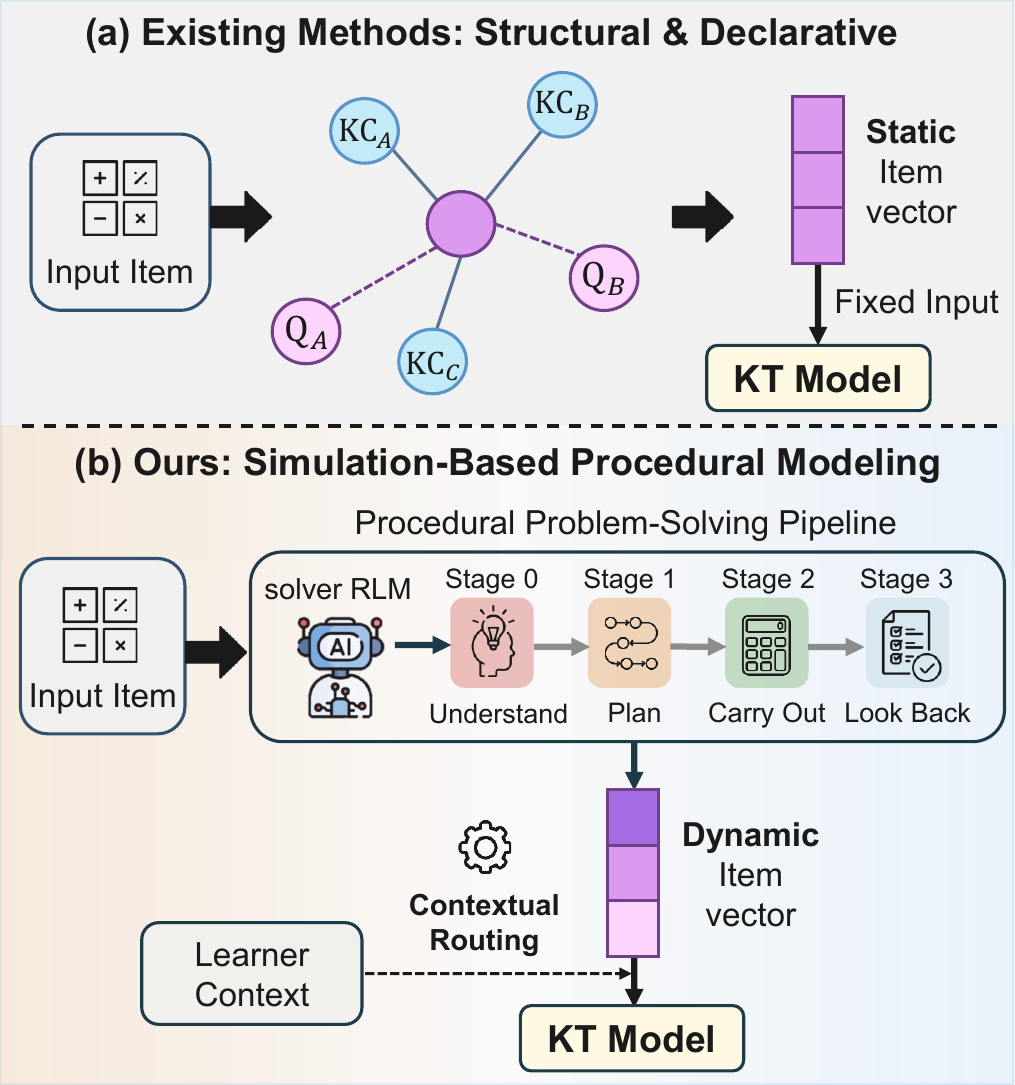}
    \caption{Comparison between conventional item modeling and BAIM. (a) Prior KT models rely on static item embeddings derived from item–KC structures. (b) BAIM instead generates procedural, context-aware item representations via Polya-based reasoning simulation.}
    \label{fig: overview}
\end{figure}

Knowledge Tracing (KT) aims to predict a learner’s future performance (i.e., whether they can correctly solve a new problem) from historical interaction data \citep{corbett1994knowledge}. A critical factor in KT is the quality of item representations, as they affect a model's ability to capture dependencies among items and to update learners' knowledge states from observed responses. 
%However, recent deep-learning-based KT models primarily focus on improving temporal prediction via sequence modeling, while leaving the representation of individual items largely underexplored 
While recent deep learning-based KT models primarily focus on improving temporal prediction through sequence modeling~\cite{huang2023towards, xu2023learning, huang2024remembering}, the representation of individual items remains largely underexplored. 
As a result, item identifiers are often mapped to randomly initialized embeddings learned solely from sparse and highly imbalanced interaction data, making it difficult to acquire robust semantic item representations \cite{krivich2025systematic}.

To address this limitation, recent work has proposed pre-trained item embedding methods that encode structural relationships between items and associated Knowledge Components (KCs) \citep{liu2020improving, wang2022perm, song2022bi, 10.1145/3638055, ozyurt2024automated}. 
However, these approaches primarily encode declarative components into static representations, overlooking the procedural dynamics of the problem-solving process. In practice, solving a problem involves multiple stages\textemdash such as interpreting the problem, setting solution strategies, and executing calculations\textemdash each reflecting distinct procedural demands, even for items associated with the same underlying concept~\cite{schoenfeld2014mathematical}. Moreover, the relative importance of these stages varies with a learner’s knowledge state and interaction history~\cite{schoenfeld1982problem}. Therefore, capturing such learner-dependent variability requires item representations that move beyond static embeddings toward adaptive modeling of procedural solution processes.

In this paper, we propose \textbf{Behavior-Aware Item Modeling (BAIM)}, a novel framework that represents items through their problem-solving processes and adapts these representations to individual learners~(Figure~\ref{fig: overview}). Grounded in Polya's four-stage problem-solving process \cite{polya1957solve} (i.e., Understanding, Planning, Carrying Out, and Looking Back), BAIM decomposes each item into structured solution stages. For each stage, BAIM leverages a reasoning language model (RLM) to derive stage-wise solution representations that capture rich latent embedding trajectories beyond surface-level item content. To adaptively leverage these stage-wise representations, BAIM introduces a context-conditioned routing mechanism that emphasizes the most informative problem-solving stage based on the learner's prior interactions. Notably, BAIM avoids auxiliary network pre-training by using one-time RLM inference and internalizes the adaptive routing mechanism directly into the KT model for unified end-to-end training.

We evaluate BAIM on the XES3G5M~\cite{liu2023xes3g5m} and NIPS34~\cite{wang2020instructions} benchmarks, where it consistently outperforms strong pretraining-based item embedding methods. In particular, BAIM exhibits clear advantages in repeated problem-solving attempts, highlighting its ability to adapt item representations to evolving learner–item interactions. Further analysis shows that leveraging the embedding trajectory yields richer and more transferable representations than using final-layer or text-only encodings. In addition, BAIM achieves faster performance gains in low-data regimes, highlighting its effectiveness under realistic educational constraints.
Our main contributions are summarized as follows:

\begin{itemize}
    \item 
    We propose BAIM, a stage-based item modeling framework grounded in Polya's problem-solving theory, representing items through structured problem-solving stages.

    \item 
    We derive stage-level representations from embedding trajectories of an RLM, capturing cognitive signals beyond surface semantics.

    \item 
    We introduce a context-conditioned routing mechanism to adaptively integrate stage-level solution representations according to the learner’s interaction history.
    
    \item 

    Extensive experiments demonstrate BAIM's robustness and adaptability in realistic settings, including repeated problem-solving attempts and low-data regimes.

\end{itemize}

\section{Related Work}

\paragraph{Item Representation learning in KT}

Recent work on item representation learning in KT has focused on enriching item representations by leveraging KCs and their relational dependencies. PEBG \cite{liu2020improving} introduced bipartite graph–based representations that explicitly encode item–KC interactions. Subsequent self-supervised approaches extended this direction by leveraging KC-anchored relational structures through diverse learning objectives, including contrastive and relation-based pretraining \cite{wang2022perm, song2022bi, 10.1145/3638055, lee2024difficulty}. More recent work explores generative approaches; KCQRL \cite{ozyurt2024automated} leverages LLM-generated step-by-step solutions to automatically annotate KCs and learn enriched item representations via contrastive objectives. 
Despite their effectiveness, they embed items as \emph{static vectors} that primarily reflect declarative knowledge or structural similarity, leaving the procedural dynamics of problem-solving largely unmodeled. 
Moreover, they typically require additional network pre-training of item representations when new items are introduced.
To address these limitations, we model item representations via procedural solution processes, capturing problem-solving dynamics beyond KC-centric structures, enabling adaptive and context-aware representations without additional pre-training as data evolves.

\label{sec:methodology}

\begin{figure*}[t]
    \centering
    
    \includegraphics[width=0.9\textwidth]{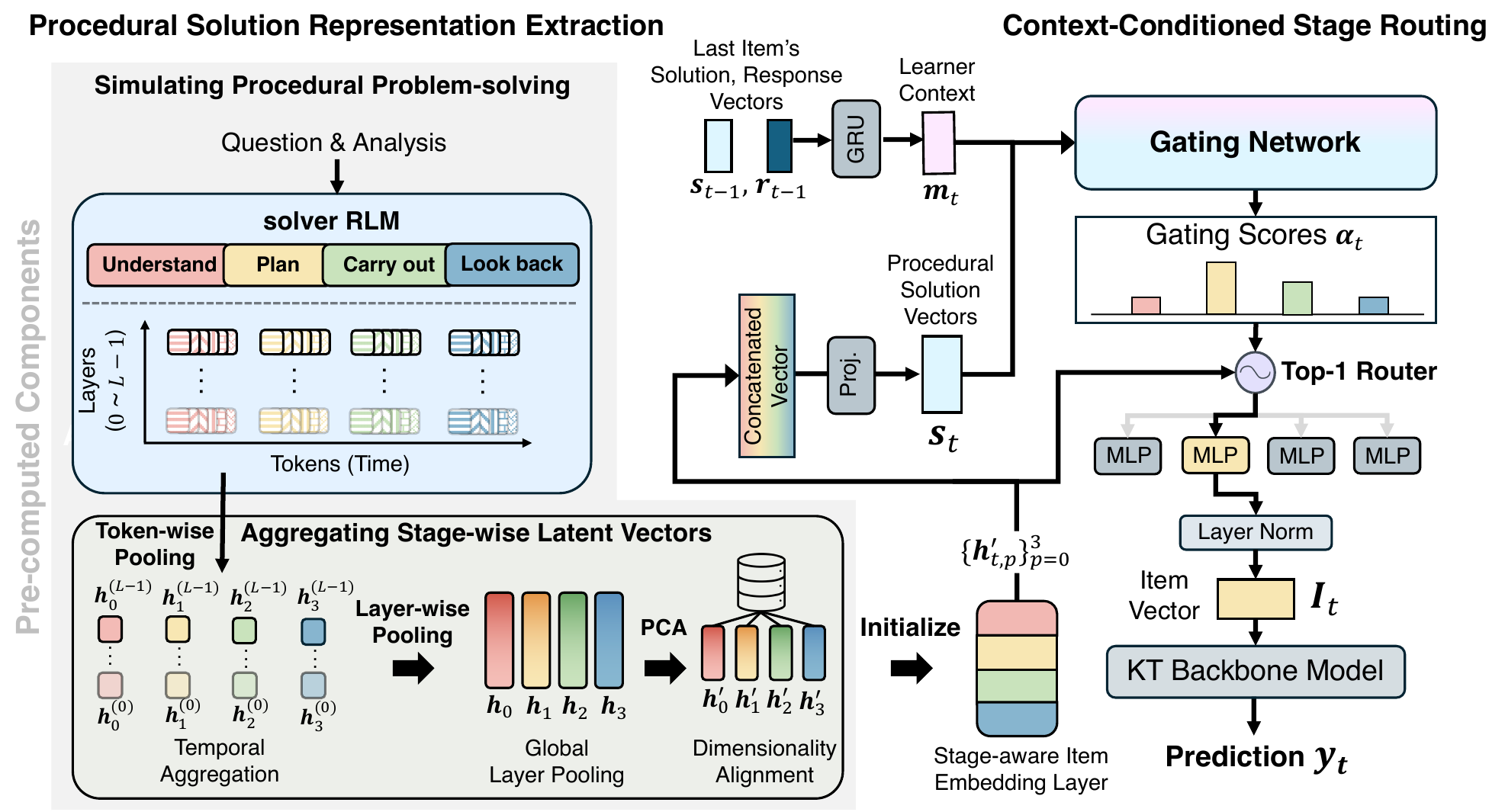}

    \caption{Overview of the proposed BAIM framework. \textbf{Left:} A solver RLM simulates Polya-based procedural problem solving for each item and extracts stage-wise latent representations. \textbf{Center:} Stage-aware procedural embeddings yield procedural solution vectors, and a learner context encoder encodes the learner’s interaction history; both serve as inputs to the gating network. \textbf{Right:} A context-conditioned stage routing module adaptively selects the most relevant problem-solving stage to produce a learner-conditioned item representation for KT.}
    \label{fig:baim_framework}
\end{figure*}

\paragraph{Deep Knowledge Tracing}

KT research has focused on developing neural architectures for modeling learners' interaction sequences.
DKT~\cite{piech2015deep} introduced LSTM-based sequence modeling, while qDKT~\cite{sonkar2020qdkt} highlighted the necessity of item-level distinctions among problems sharing the same KC.
With the adoption of Transformer~\cite{NIPS2017_3f5ee243} architectures, AKT~\cite{ghosh2020context} further advanced the field by combining monotonic decay attention for sequence modeling with Rasch-based embeddings for enhanced item representation. Subsequent work has emphasized simplicity and robustness in model design; for example, simpleKT \cite{liu2023simplekt} and sparseKT~\cite{huang2023towards} achieved competitive performance by emphasizing architectural simplicity and sparse attention mechanisms, respectively. In parallel, efforts to improve interpretability have led to cognitively grounded designs such as QIKT~\cite{chen2023improving}, which adopts item-centric cognitive representations leveraging associated KCs as auxiliary information, and incorporates an item response theory-based prediction layer. To evaluate our item modeling approach, we integrate BAIM into the item representation components of these KT backbones, while preserving their original sequence modeling and prediction mechanisms.

\section{Problem Statement}
\label{sec:problem_statement}

Following prior work \cite{sonkar2020qdkt}, we adopt the standard \emph{item-level} formulation of KT, where the objective is to predict a learner's response to a given item at time step $t$ based on the learner’s historical interaction sequence. A learner's interaction history is represented as a temporal sequence
$X = \{x_0, x_1, \dots, x_{t-1}\}$,
where the $j$-th interaction $x_j$ is defined as a 2-tuple
$x_j = (I_j, r_j)$.
Here, $I_j$ denotes a unique item identifier,
and $r_j \in \{0,1\}$ indicates whether the learner answered the question of the item correctly. Given the historical interactions $\{x_j\}_{j=0}^{t-1}$ and the current item $I_t$, a KT model estimates the probability that the learner answers the item correctly:
\begin{equation}
y_t = P(r_t = 1 \mid I_t, \{x_j\}_{j=0}^{t-1}).
\end{equation}

The model is trained by minimizing the binary cross-entropy loss
between the predicted probability $y_t$ and the observed response $r_t$:
\begin{equation}
\mathcal{L}_{\text{KT}}
= - \sum_{t}
\left[
r_t \log y_t + (1 - r_t) \log (1 - y_t)
\right].
\end{equation}

In this item-centric formulation, the item identifier $I_j$ serves as the primary modeling unit, and all predictions and losses are defined with respect to item responses. KCs, when available, are treated as contextual information rather than independent prediction targets.

\section{Behavior-Aware Item Modeling (BAIM)}

Existing item representation learning approaches primarily rely on KC tags, which, from the ACT-R perspective, represent \emph{declarative knowledge}~\citep{anderson1996act}. However, this focus overlooks \emph{procedural knowledge}—the process and capability involved in solving problems. To address this limitation, we propose BAIM, which captures both aspects by explicitly modeling the act of solving itself. BAIM operates by integrating procedural solution representations with item- and learner-conditioned contextual signals.
First, it extracts procedural solution representations for each item by decomposing the solution process into structured problem-solving stages following Polya’s framework, including \emph{Understand}, \emph{Plan}, \emph{Carry Out}, and \emph{Look Back}. Second, BAIM introduces a \emph{Context-Conditioned Stage Routing} mechanism that adaptively determines which problem-solving stage should be emphasized for a given item, conditioned on the learner’s interaction context. Through this routing mechanism, BAIM aligns item-level procedural characteristics with the learner’s context, enabling more fine-grained personalized diagnosis and prediction. The overall process is illustrated in Figure~\ref{fig:baim_framework}.

\subsection{Procedural Solution Representation Extraction}

\begin{figure}[t]
\centering
\small

% --- Image ---
\includegraphics[width=\columnwidth]{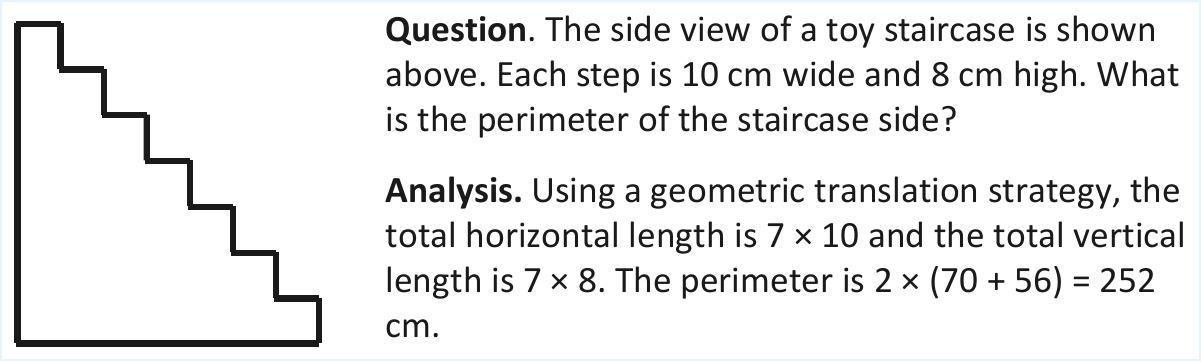}
\vspace{0.6em}
\begin{tabular}{
  >{\centering\arraybackslash}m{0.24\columnwidth}
  m{0.65\columnwidth}
}
\toprule
\textbf{($p$) Stage} & \textbf{Stage-wise Solution Description} \\
\midrule
\rowcolor{red!10}
\textbf{\shortstack{(0) Understand}} &
Each step is 10\,cm wide (tread) and 8\,cm high (riser). There are 7 steps. The goal is to calculate the perimeter of the staircase side view. \\
\rowcolor{yellow!15}
\textbf{\shortstack{(1) Plan}} &
Use the translation method: total horizontal segments $= 7 \times 10$, total vertical segments $= 7 \times 8$. The perimeter is $2 \times$ (total horizontal $+$ total vertical). \\
\rowcolor{green!10}
\textbf{\shortstack{(2) Carry Out}} &
$7 \times 10 = 70$; $7 \times 8 = 56$; $70 + 56 = 126$; $126 \times 2 = 252$. \\
\rowcolor{cyan!10}
\textbf{\shortstack{(3) Look Back}} &
The perimeter is 252\,cm, which matches the analysis and accounts for all outer edges of the staircase. \\
\bottomrule
\end{tabular}
\caption{Example of a staircase-shaped geometry problem  from the XES3G5M dataset and its stage-wise solution description.}
\label{fig:staircase_onecol}
\end{figure}

\paragraph{Simulating Procedural Problem-solving.} 
The first step of BAIM simulates the problem-solving process for the given item.
We employ an RLM as a solver that generates a structured problem-solving process according to Polya’s four-stage framework, as illustrated in Figure~\ref{fig:staircase_onecol}. Given an item and its accompanying analysis, the solver RLM generates a problem-solving process together with a structured output that delineates four problem-solving stages after internal reasoning. Each stage $p \in \{0,1,2,3\}$ corresponds to a contiguous token span $(T_{s,p}, T_{e,p})$. During this process, the RLM naturally generates token- and layer-wise latent vectors, which form the basis for subsequent stage-wise aggregation. Details of the prompting strategy are provided in Appendix~\ref{sec:solver_prompt}.

\paragraph{Aggregating Stage-wise Latent Vectors.}

We construct procedural representations for KT by aggregating embedding trajectories produced by an RLM at different granularities.
Specifically, we first perform temporal aggregation by aligning RLM's token-level hidden states along the generation timeline with predefined problem-solving stages, motivated by prior work that demonstrates the effectiveness of aggregating internal hidden states over the generated sequence for tasks such as out-of-distribution detection and self-evaluation~\cite{wang2024embedding, wang2024latent}.
We then apply global pooling over the resulting embedding trajectories to integrate procedural signals captured across the full trajectory.

\paragraph{(a) Temporal Aggregation.}
For each problem-solving stage $p$ and each transformer layer $l$, we aggregate token-level hidden states within the stage-specific token span $[T_{s,p}, T_{e,p}]$ using mean pooling.  
This operation summarizes the latent state of the solver corresponding to stage $p$ at depth $l$:
\vspace{-10pt}

\begin{small}
\begin{equation}
\mathbf{h}_p^{(l)} =
\frac{1}{T_{e,p} - T_{s,p} + 1}
\sum_{k=T_{s,p}}^{T_{e,p}} \mathbf{h}_k^{(l)},
\quad
\mathbf{h}_k^{(l)} \in \mathbb{R}^{D_{\text{solver}}}.
\end{equation}
\end{small}
where $\mathbf{h}_p^{(l)}$ denotes a latent vector associated with stage $p$ at layer $l$ and is treated as an intermediate quantity rather than a final representation.

\paragraph{(b) Global Layer Pooling.}
The stage-wise latent vectors obtained above still retain layer-specific variations. Latent vectors extracted from different layers capture complementary aspects of the solution process, reflecting how information is progressively transformed across the model depth. Recent studies suggest that relying on a single or shallow subset of layers may overlook useful procedural information encoded in intermediate representations \citep{tang2024pooling, skean2025layer}. Accordingly, we perform entire layer pooling by averaging across all $L$ transformer layers: 
\begin{equation}
\mathbf{h}_p = \frac{1}{L} \sum_{l=0}^{L-1} \mathbf{h}_p^{(l)}. 
\end{equation} 
This operation yields a unified stage-level latent summary that integrates information distributed throughout the entire model.

\paragraph{(c) Dimensionality Alignment.}
Since the dimensionality of $\mathbf{h}_p$ exceeds that of standard KT backbones, we apply PCA~\cite{bishop2006pattern} to reduce it to 768 dimensions for compatibility with prior work~\cite{ozyurt2024automated}.
The resulting vectors are used as the stage-level representations for item $I_t$, denoted as $\{\mathbf{h}'_{p}\}_{p=0}^{3}$. These representations are precomputed for all items and used to initialize
the item embedding layer of the KT model, providing each item with stage-aware procedural priors. After initialization, the embedding corresponding to item $I_t$ and stage $p$, denoted as $\mathbf{h}'_{t,p}$, is optimized jointly with the training objective rather than being frozen.

\subsection{Context-Conditioned Stage Routing}

Not all problem-solving stages are equally informative in practice.
Prior work in mathematical problem solving and cognitive load theory suggests that the cognitive demands and diagnostic relevance of each stage vary across problems and depend on the items' and learners' context \citep{sweller1988cognitive, schoenfeld2014mathematical}. Therefore, we propose a \emph{Context-Conditioned Stage Routing} mechanism that adaptively adjusts the emphasis placed on different problem-solving stages based on the learner context. All architectural specifications, including dimensionalities and parameter
configurations, are provided in Appendix~\ref{sec:appendix_baim_architecture}.

\paragraph{Procedural Solution Encoding.}
Given the stage-level representations
$\{\mathbf{h}'_{t,p}\}_{p=0}^{3}$ for item $I_t$,
we aggregate them to form a single \emph{procedural solution representation}:
\begin{equation}
\mathbf{s}_t =
f_{\text{proj}}
\left(
\mathbf{h}'_{t,0} \oplus
\mathbf{h}'_{t,1} \oplus
\mathbf{h}'_{t,2} \oplus
\mathbf{h}'_{t,3}
\right),
\end{equation}
where $\oplus$ denotes concatenation and $f_{\text{proj}}(\cdot)$ denotes a
learnable projection function. The resulting vector $\mathbf{s}_t$ provides a unified procedural encoding of
item $I_t$ and serves as the item-side input to the context-conditioned stage
routing mechanism.

\paragraph{Learner Context Encoding.}
To condition stage routing on the learner’s historical interaction patterns,
we encode the learner’s context into a latent context vector $\mathbf{m}_t$.
Given the learner’s interactions history $\{(I_j, r_j)\}_{j=0}^{t-1}$, the learner context is updated using a Gated Recurrent Unit (GRU) \cite{chung2014empirical}:
\begin{equation}
\mathbf{m}_t =
\mathrm{GRU}
\left(
\mathbf{m}_{t-1},
\mathbf{W}_{\text{in}}
\left[
\mathbf{s}_{t-1} \oplus \mathbf{r}_{t-1}
\right]
\right),
\end{equation}

This formulation provides a contextual signal that conditions the subsequent stage routing process.

\paragraph{Top-1 Stage Routing Mechanism.}
Given the procedural solution vector $\mathbf{s}_t$ and the learner context vector $\mathbf{m}_t$,
the routing module computes gating scores
$\boldsymbol{\alpha}_t \in \mathbb{R}^4$
by jointly conditioning on both signals:
\begin{equation}
\boldsymbol{\alpha}_t =
\mathbf{W}_{\text{gate}}
\left[
\mathbf{s}_t \oplus \mathbf{m}_t
\right]
+
\mathbf{b}_{\text{gate}}.
\end{equation}
Conditioned on the joint item–learner context, we adopt a Top-1 routing strategy,
selecting the stage with the highest gating score (i.e., $k^* = \arg\max \boldsymbol{\alpha}_t$), thereby assigning different
importance to problem-solving stages.

\paragraph{Stage-Specific Expert Transformation.}
The selected stage-level representation $\mathbf{h}'_{t,k^*}$ is transformed by
a stage-specific expert to produce the final context-conditioned item
representation: 
\begin{equation}
\mathbf{I}_t =
\mathrm{LayerNorm}
\left(
\mathrm{MLP}_{k^*}
\left(
\mathbf{h}'_{t,k^*}
\right)
\right).
\end{equation}
The resulting vector $\mathbf{I}_t$ constitutes the final context-conditioned item representation and is consumed by the downstream KT backbone.

\subsection{Objective Function}

BAIM is trained under the standard item-level KT objective
defined in Section~\ref{sec:problem_statement}.
To prevent routing collapse, we additionally apply a load-balancing regularization
term following Switch Transformers~\cite{fedus2022switch}. Specifically, given the gating scores $\boldsymbol{\alpha}_t$,
we compute gating probabilities
$\mathbf{p}_t = \mathrm{Softmax}(\boldsymbol{\alpha}_t)$.
Let $\bar{p}_j$ denote the batch-averaged probability of selecting stage $j$, load-balancing loss is defined as:
\begin{equation}
\mathcal{L}_{\text{LB}} =
\sum_{j=0}^{3}
\left(
\bar{p}_j - \frac{1}{4}
\right)^2.
\end{equation}

We define the final training objective as:
\begin{equation}
\mathcal{L}_{\text{total}}
=
\mathcal{L}_{\text{KT}} + \lambda \mathcal{L}_{\text{LB}}.
\end{equation}
\section{Experimental Setup}
% 상원 수정
\paragraph{Datasets}
We evaluate BAIM on two real-world mathematical KT benchmarks
that provide publicly available item-level metadata:
\textbf{XES3G5M}~\cite{liu2023xes3g5m}\footnote{\url{https://github.com/ai4ed/XES3G5M.git}}
and \textbf{NIPS34}~\cite{wang2020instructions}.\footnote{\url{https://www.eedi.com/research}}
XES3G5M serves as our primary large-scale benchmark,
and we additionally evaluate on NIPS34 to validate its robustness.
Dataset statistics and metadata characteristics are summarized in
Table~\ref{tab:data_statistics}.
Due to differences in item and analysis metadata formats,
we apply dataset-specific preprocessing to enable reasoning-based item modeling.
Further details are provided in Appendix~\ref{sec:appendix_preprocessing}.

\paragraph{Solver RLM}
We use Qwen3-VL-32B-Thinking \cite{bai2025qwen3vltechnicalreport} as the solver RLM to generate procedural solution representations for items. The solver is used offline to extract stage-level representation for each item. The model comprises $L = 65$ transformer layers with a hidden dimensionality of
$D_{\text{solver}} = 5{,}120$. 
% and does not participate in KT model training.

\begin{table}[t]
\centering
\small
\scalebox{0.92}{
\begin{tabular}{lcc}
\toprule
\multirow{2}{*}{\centering \textbf{Attribute}}
& \multicolumn{2}{c}{\textbf{Datasets}} \\
\cmidrule(lr){2-3}
& \textbf{XES3G5M} 
& \textbf{NIPS34} \\
\midrule
\# Students        & 18.07K        & 4.92K \\
\# Questions        & 7.65K         & 948 \\
\# KCs          & 865           & 57 \\
\# Interactions         & 5.55M         & 1.38M \\
Question Meta  & Text + Image (CN) & Image-only (EN) \\
Analysis Meta  & Provided      & Not provided \\
\bottomrule
\end{tabular}
}
\caption{
Dataset statistics and metadata availability.
Question Meta indicates the original format of question content,
and Analysis Meta denotes whether analytical explanations are provided.
}
\label{tab:data_statistics}
\end{table}

% 순서 변경
\paragraph{Item Representation Baselines}
We compare BAIM against two well-established item representation baselines while keeping the underlying KT backbone architectures fixed.
The \textbf{Default} setting uses randomly initialized item- and KC-level embeddings that are trained end-to-end together with each backbone.
For \textbf{pre-trained baselines}, we reproduce 768-dimensional item embeddings using the official implementations of \texttt{PEBG}\footnote{\url{https://github.com/ApexEDM/PEBG.git}} and \texttt{KCQRL}.\footnote{\url{https://github.com/oezyurty/KCQRL.git}}
Implementation details and reproduction-specific adaptations are provided in Appendix~\ref{sec:appendix_baselines}.

\paragraph{Backbone KT Models}
We select five representative KT backbones—\textbf{AKT}, \textbf{qDKT}, \textbf{QIKT}, \textbf{simpleKT}, and \textbf{sparseKT}—which achieved strong performance in the KCQRL benchmark, and adapt them to our framework for evaluation. For each backbone, we replace the \textit{Item Representation Module} with BAIM while keeping all other architectural components unchanged. Backbone-specific integration details are in Appendix~\ref{sec:appendix_baim_backbones}.

\paragraph{Evaluation}
We evaluate the effectiveness of item representations on each dataset by measuring their impact on KT performance using AUC. Results for which the mean or standard deviation is reported are obtained via 5-fold cross-validation, with the random seed fixed to 42 for reproducibility.

\paragraph{Training Details}
All models are trained with a batch size of 128, an embedding size of 256, and a dropout rate of 0.1. We fix the learning rate to $1 \times 10^{-4}$ and adopt default hyperparameter settings from the \texttt{pyKT} library.\footnote{\url{https://pykt.org}} We set the loss weighting coefficient $\lambda$ to $0.01$, following the setting used in Switch Transformers~\cite{fedus2022switch}.

\begin{table*}[t]
\centering
\small
\setlength{\tabcolsep}{3pt} 
\begin{tabular}{l|ccccc}
\toprule
\multirow{2}{*}{\diagbox[width=7em]{Embed}{Model}} & \multicolumn{5}{c}{Knowledge Tracing Backbone Architecture} \\
\cmidrule(l){2-6}
 & AKT & qDKT & QIKT & simpleKT & sparseKT \\
\midrule

\multicolumn{6}{l}{\textbf{XES3G5M}}\\
\midrule
Default 
& 81.56 {\scriptsize $\pm$ 0.06}
& 81.69 {\scriptsize $\pm$ 0.04}
& 81.67 {\scriptsize $\pm$ 0.01}
& 81.26 {\scriptsize $\pm$ 0.01}
& 80.37 {\scriptsize $\pm$ 0.04} \\
PEBG   
& 82.79 {\scriptsize $\pm$ 0.04} (+1.23) 
& 82.16 {\scriptsize $\pm$ 0.04} (+0.47)
& 82.00 {\scriptsize $\pm$ 0.02} (+0.33)
& 82.51 {\scriptsize $\pm$ 0.02} (+1.25)
& 82.63 {\scriptsize $\pm$ 0.07} (+2.26) \\
KCQRL   
& 82.67 {\scriptsize $\pm$ 0.03} (+1.11)
& 81.94 {\scriptsize $\pm$ 0.03} (+0.25)
& 81.85 {\scriptsize $\pm$ 0.02} (+0.18)
& 82.48 {\scriptsize $\pm$ 0.02} (+1.22)
& 82.61 {\scriptsize $\pm$ 0.11} (+2.24) \\
\textbf{BAIM} (Ours) 
& \textbf{83.00} {\scriptsize $\pm$ 0.04} (\textbf{+1.44})
& \textbf{82.43} {\scriptsize $\pm$ 0.02} (\textbf{+0.74})
& \textbf{82.17} {\scriptsize $\pm$ 0.05} (\textbf{+0.50})
& \textbf{82.84} {\scriptsize $\pm$ 0.01} (\textbf{+1.58})
& \textbf{83.21} {\scriptsize $\pm$ 0.10} (\textbf{+2.84}) \\

\midrule
\multicolumn{6}{l}{\textbf{NIPS34}}\\
\midrule
Default 
& 79.89 {\scriptsize $\pm$ 0.07}
& 79.24 {\scriptsize $\pm$ 0.08}
& 79.95 {\scriptsize $\pm$ 0.07}
& 79.90 {\scriptsize $\pm$ 0.01}
& 79.30 {\scriptsize $\pm$ 0.08} \\
PEBG    
& 80.10 {\scriptsize $\pm$ 0.02} (+0.21)
& 80.10 {\scriptsize $\pm$ 0.03} (+0.86)
& 80.15 {\scriptsize $\pm$ 0.03} (+0.20)
& 79.96 {\scriptsize $\pm$ 0.01} (+0.06)
& 80.21 {\scriptsize $\pm$ 0.17} (+0.91) \\
\textbf{BAIM} (Ours) 
& \textbf{80.16} {\scriptsize $\pm$ 0.04} (\textbf{+0.27})
& \textbf{80.13} {\scriptsize $\pm$ 0.03} (\textbf{+0.89})
& \textbf{80.18} {\scriptsize $\pm$ 0.04} (\textbf{+0.23})
& \textbf{80.02} {\scriptsize $\pm$ 0.03} (\textbf{+0.12})
& \textbf{80.36} {\scriptsize $\pm$ 0.12} (\textbf{+1.06}) \\
\bottomrule
\end{tabular}
\caption{
Performance comparison (mean $\pm$ std) of KT backbones with different item representation methods.
Values in parentheses denote absolute improvements over the Default setting; The best-performing method in terms of mean AUC is highlighted in \textbf{bold}.
}
\label{tab:main_result}
\end{table*}

\section{Results}

% 상원 수정
\paragraph{Main Results}
To assess the effectiveness of BAIM, we compare BAIM against strong item representation baselines across five representative KT backbones on XES3G5M and NIPS34, demonstrating clear improvements in KT performance.
As shown in Table~\ref{tab:main_result}, BAIM achieves the highest AUC across all KT architectures on the XES3G5M dataset.
Compared to both randomly initialized embeddings (\texttt{Default}) and pre-trained item representations (\texttt{PEBG} and \texttt{KCQRL}), BAIM yields performance gains across diverse backbone designs.
In particular, BAIM achieves the largest gain on \texttt{sparseKT}, improving AUC from 80.37 to 83.21, suggesting that behavior-aware item representations are especially effective when combined with sparse attention mechanisms.
Notably, the advantages of BAIM extend beyond large-scale, text-rich settings.
On the smaller-scale NIPS34 dataset, which differs substantially in both scale and metadata format, BAIM continues to outperform both the \texttt{Default} and \texttt{PEBG} baselines across all KT backbones.
Overall, these results indicate that BAIM generalizes well across datasets with heterogeneous characteristics, highlighting the robustness of behavior-aware item modeling beyond specific data conditions.

\paragraph{Stage-Level Dynamics under Repeated Interactions}
\begin{figure}[t]
    \centering
    \includegraphics[width=0.9\linewidth]{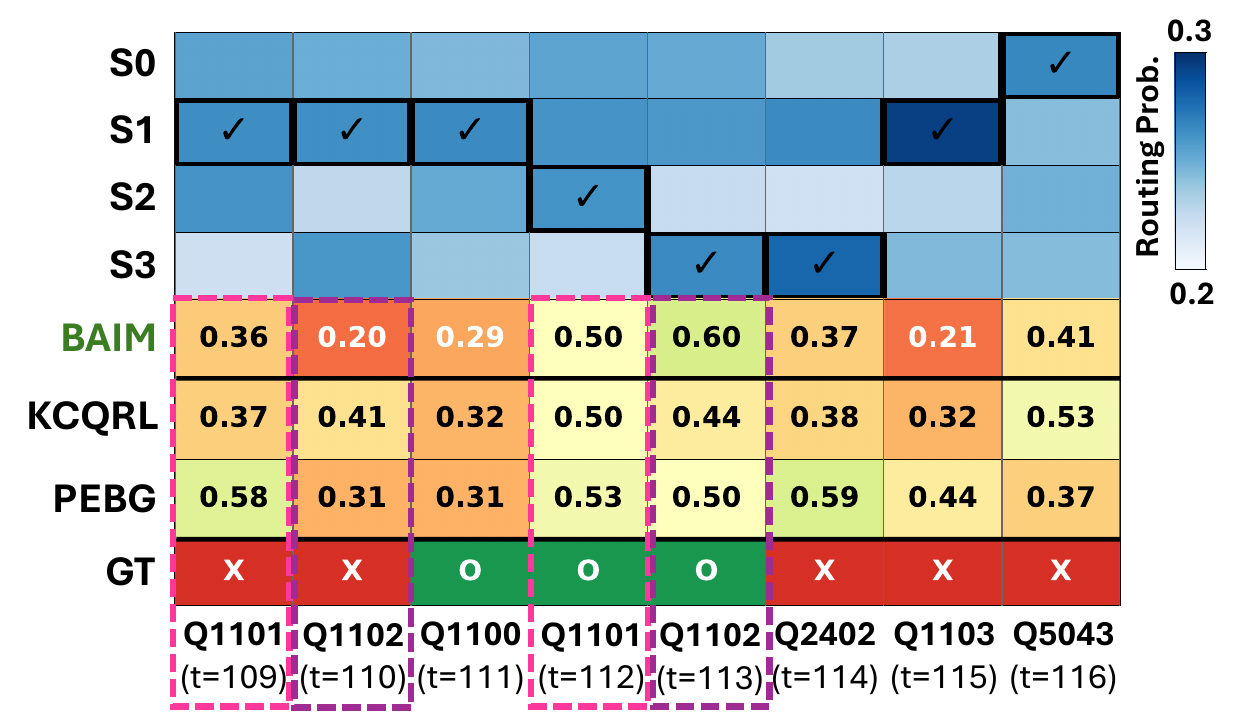}
    \caption{Visualization of BAIM's stage-level dynamic routing behavior and its impact on prediction outcomes under repeated interactions. \textbf{Rows 0--3}: routing weights over problem-solving stages; \textbf{Rows 4--6}: prediction outcomes of BAIM and baseline methods; \textbf{Row 7}: ground-truth (GT) student performances.}

    \label{fig:case_study}
\end{figure}

\begin{figure}[t]
    \centering
    \includegraphics[width=0.82\linewidth]{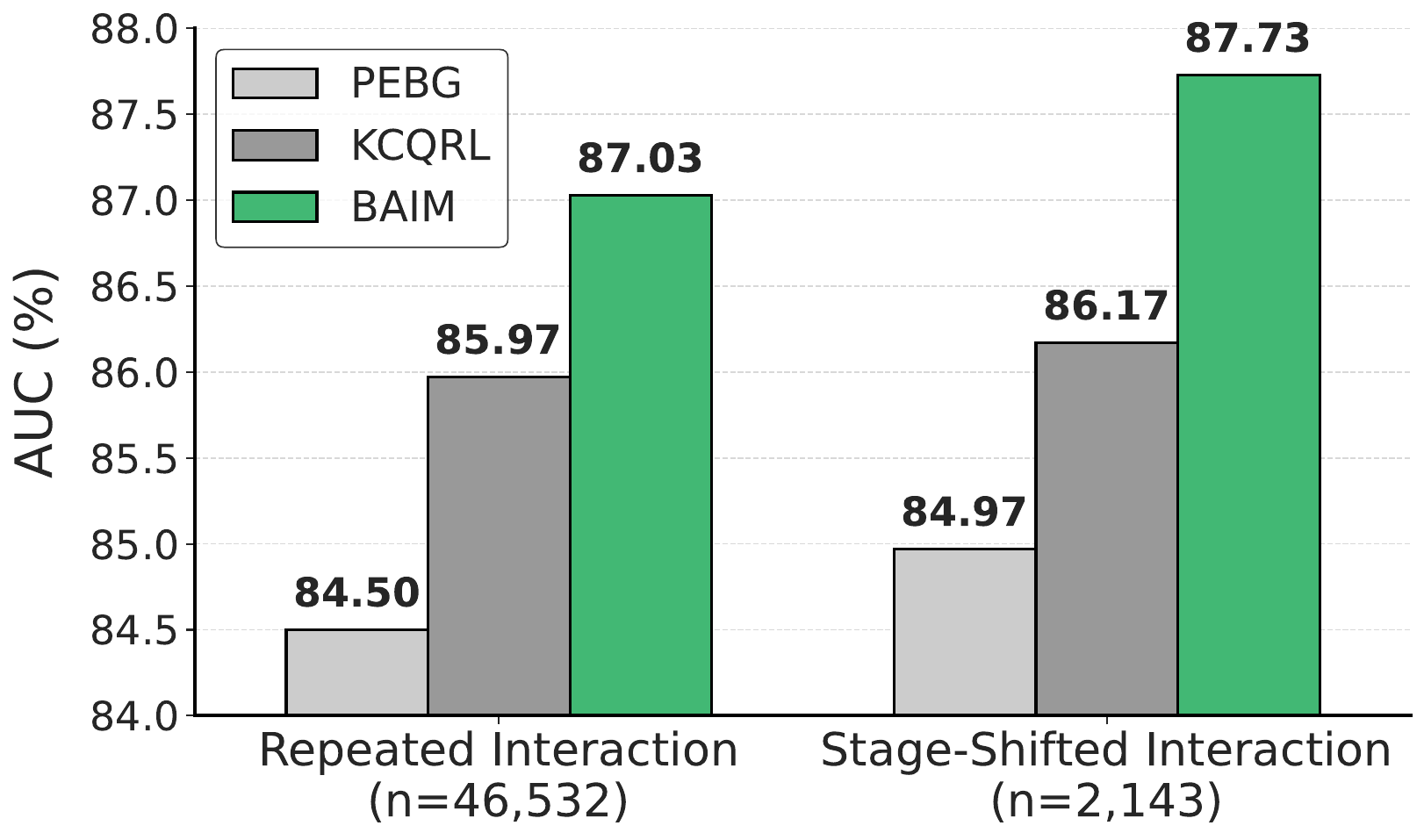}
    \caption{AUC comparison across repeated student interactions. Adaptive routing enables procedural focus shifts in BAIM.
    }
    \label{fig:auc_performance}
\end{figure}

To evaluate the adaptability of item representation methods, we analyze a sequence of learner-item interactions from the XES3G5M dataset using a sparseKT model trained on fold 0. Figure~\ref{fig:case_study} visualizes BAIM’s routing probabilities across problem-solving stages over time, alongside prediction outcomes from baseline methods. BAIM dynamically adjusts its routing focus to emphasize different solution stages as the learner’s interaction context evolves, reflecting changes in procedural demand. This adaptive behavior is particularly evident for items Q1101 and Q1102, which are reattempted multiple times by the same learner. In contrast, baseline methods rely on fixed item representations, which may limit their adaptability and potentially account for their inaccurate predictive performance.

Quantitative results in Figure~\ref{fig:auc_performance} further support the effectiveness of adaptive routing. Among $46{,}532$ repeated interactions, BAIM changes its routing decision in $2{,}143$ cases (stage-shifted subset) where adaptive behavior is explicitly triggered. BAIM outperforms the strongest baseline by 1.06 AUC on all repeated interactions, with the margin increasing to 1.56 AUC on the stage-shifted subset, highlighting the effectiveness of dynamically adapting stage-wise item representations.

\section{Analysis}

\begin{figure*}[t]
    \centering
    \includegraphics[width=1\textwidth]{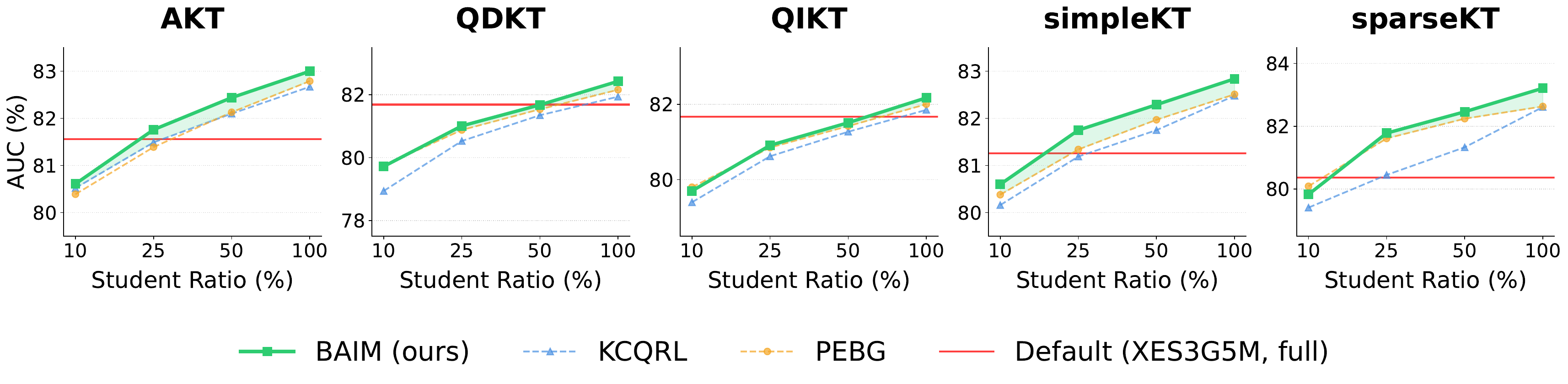}
    \caption{
    AUC (\%, mean) of five KT backbones trained with varying numbers of students on XES3G5M.
    The red horizontal line indicates the performance of the Default representation
    trained on the full dataset (100\%).
    }
    \label{fig:subset_performance}
\end{figure*}

\paragraph{Impact of Routing Strategy}

\begin{figure}[t]
    \centering
    \includegraphics[width=\columnwidth]{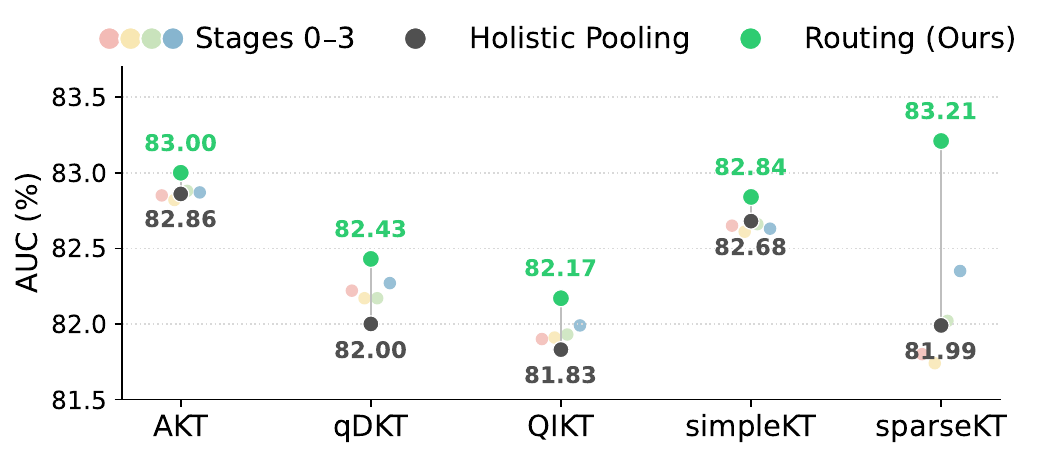}
    \caption{
    Performance comparison of aggregation strategies across multiple KT backbones on XES3G5M, highlighting the consistent advantage of BAIM’s routing-based aggregation over fixed schemes.
    }
    \label{fig:routing_vs_fixed}
\end{figure}

We compare an adaptive routing strategy in BAIM with fixed aggregation strategies, including
(i) selecting a single solution stage (Stage 0–3) and
(ii)  holistic pooling over the full solution trajectory without stage decomposition.
Figure~\ref{fig:routing_vs_fixed} shows that our adaptive routing strategy outperforms all single-stage models across all KT backbones, indicating that the most informative stage varies with the problem characteristics. Compared with holistic pooling, adaptive routing also yields consistently better performance, supporting the importance of explicit stage decomposition rather than relying on a single global representation.

\paragraph{Impact of Representation Extraction Strategies}
\label{sec:ablation_pooling}

\begin{figure}[t]
    \centering
    \includegraphics[width=\columnwidth]{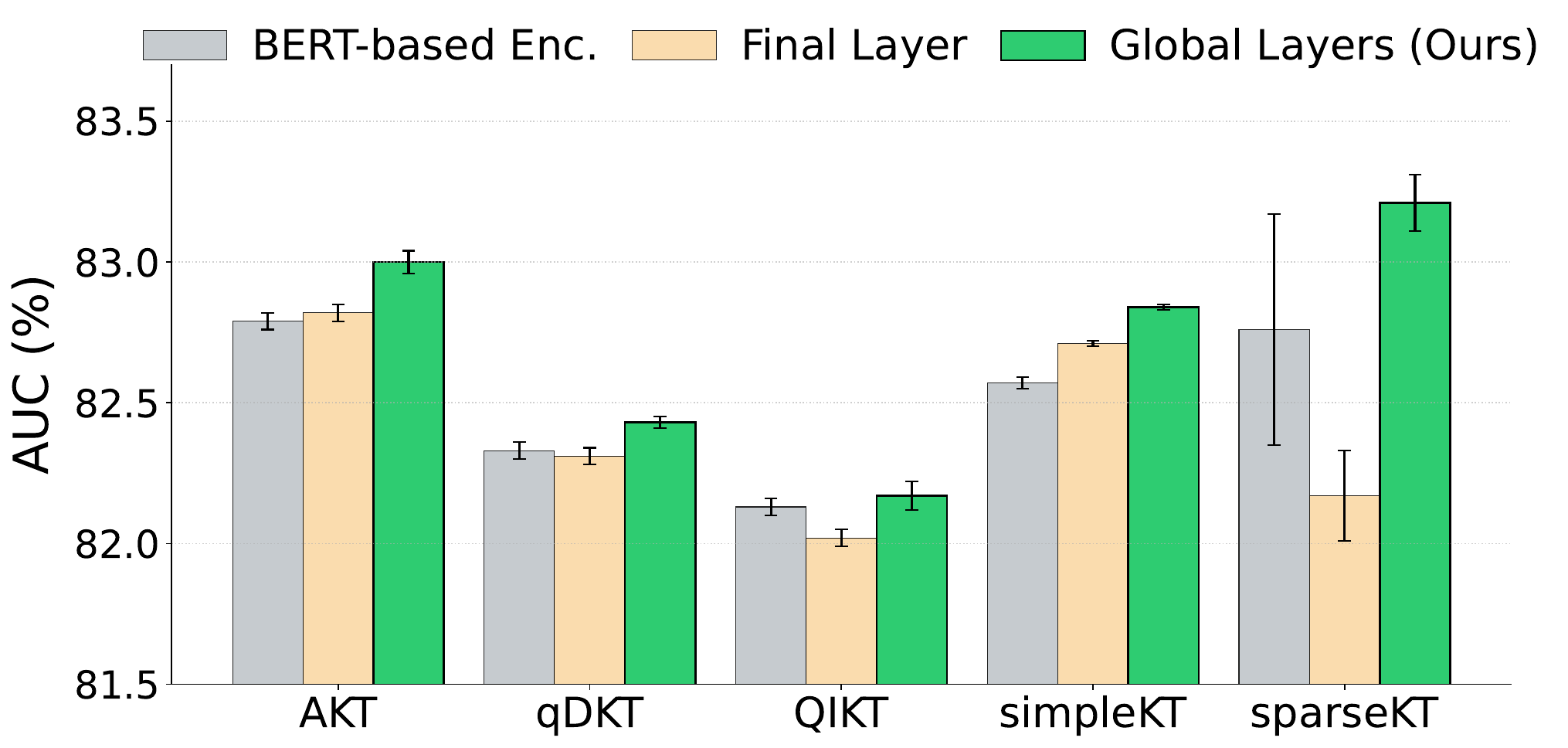}
    \caption{Impact of different representation extraction strategies. Leveraging the embedding trajectory of the solver RLM yields more effective item representations.
    }
    \label{fig:ablation_layer_pooling}
\end{figure}

We investigate the effect of different representation extraction strategies from the RLM solution process on downstream KT performance. All components of BAIM are kept fixed, and we vary only the aggregation strategy of hidden states across RLM layers or, alternatively, over generated solution texts to form stage-wise solution representations. Figure~\ref{fig:ablation_layer_pooling} demonstrates that leveraging information from the full depth of the RLM consistently outperforms representations derived from a single layer or sentence-level encoding across all KT backbones. Specifically, global layer pooling, which aggregates representations across the entire embedding trajectory ($l=[0,64]$), achieves the strongest performance, outperforming both the final-layer ($l=64$) and BERT-based encoding. In contrast, relying solely on the final layer leads to lower performance, suggesting that single-layer representations struggle to capture the diverse procedural and semantic information distributed throughout the depth of the RLM, consistent with recent findings on the benefits of intermediate and aggregated layer representations~\cite{skean2025layer}. 

\paragraph{Sample Efficiency}

To analyze BAIM's robustness under limited training data, we vary the number of training students by randomly subsampling the XES3G5M dataset at ratios of 10\%, 25\%, 50\%, and 100\%. All models are trained from scratch on each subset and evaluated on the same test set. As detailed in Figure~\ref{fig:subset_performance}, BAIM achieves strong performance even with substantially fewer training students.
Notably, for AKT, simpleKT, and sparseKT, BAIM trained with only 25\% of the students already outperforms the \texttt{Default} model trained on the full dataset, demonstrating pronounced sample efficiency. Starting from the 25\% ratio, BAIM consistently surpasses other item representation methods across all backbone architectures.

\section{Conclusion}

We propose BAIM, a behavior-aware item modeling framework for KT that represents items via structured problem-solving. By deriving stage-level representations from embedding trajectories of an RLM, BAIM captures rich procedural signals without auxiliary network pre-training. Moreover, BAIM adaptively integrates these representations via context-conditioned routing, allowing different problem-solving stages to be emphasized based on both procedural solution dynamics and learner histories. We demonstrate the effectiveness of BAIM across five representative KT backbones and two real-world math learning datasets, where it consistently improves prediction performance, especially under repeated learner–item interactions.

\section*{Acknowledgments}

This research was supported by the Culture, Sports and Tourism R\&D Program through the Korea Creative Content Agency grant funded by the Ministry of Culture, Sports and Tourism in 2025 (Project Name: Development of an AI-Based Korean Diagnostic System for Efficient Korean Speaking Learning by Foreigners, Project Number: RS-2025-02413038, Contribution Rate: 45\%); by the IITP(Institute of Information \& Communications Technology Planning \& Evaluation)-ITRC(Information Technology Research Center) grant funded by the Korea government(Ministry of Science and ICT) (IITP-2026-RS-2024-00437866, Contribution Rate: 45\%); and by Institute of Information \& Communications Technology Planning \& Evaluation (IITP) grant funded by the Korea government(MSIT) (No.RS-2019-II191906, Artificial Intelligence Graduate School Program(POSTECH), Contribution Rate: 10\%). 

We sincerely thank Deokhyung Kang and Chiyeong Heo for valuable discussions.

\section*{Limitations}

While BAIM effectively captures procedural item representations and consistently improves KT performance across diverse settings, several limitations remain.
First, BAIM is designed and evaluated primarily in the context of mathematical problem solving. Extending the framework to other educational domains, such as programming tasks or second language learning, may require domain-specific adaptations to the stage formulation and procedural modeling.
Second, our empirical evaluation is restricted to KT benchmarks that provide publicly available item content. As a result, we do not include several large and widely used datasets, such as ASSISTments, where item metadata are not publicly released, limiting direct comparison on these benchmarks.

\section*{Ethical Statement}

This work investigates item representation learning and does not involve ethical issues. All experiments are conducted using publicly accessible datasets. Specifically, the XES3G5M dataset and pyKT library are used under the MIT License, and the NIPS34 dataset is utilized in accordance with its official Terms of Service. The use of these datasets is strictly limited to academic research, which is consistent with their intended purpose and access conditions. GitHub Copilot was used to assist with code generation, and ChatGPT was used to support writing and language refinement. All research contributions are solely attributable to the authors.

\bibliography{custom}
\appendix

\appendix
\label{sec:appendix}

\section{Architectural Details of BAIM}
\label{sec:appendix_baim_architecture}

\begin{table}[t]
\centering
\small
\begin{tabularx}{\columnwidth}{l X}
\toprule
\textbf{Notation} & \textbf{Shape} \\
\midrule
$D_{\text{input}}$ & $768$ \\
$D_{\text{kt}}$ & $256$ \\
$D_{\text{history}}$ & $64$ \\
$f_{\text{proj}}(\cdot)$
& Linear$(4D_{\text{input}} \rightarrow D_{\text{kt}})$ + ReLU + Dropout \\
$\mathbf{W}_{\text{in}}$ & $\in \mathbb{R}^{D_{\text{history}} \times (2D_{\text{kt}})}$ \\
$\mathbf{W}_{\text{gate}}$ & $\in \mathbb{R}^{4 \times (D_{\text{kt}} + D_{\text{history}})}$ \\
$\mathbf{b}_{\text{gate}}$ & $\in \mathbb{R}^{4}$ \\
$\mathbf{r}_{t-1}$ & $\in \mathbb{R}^{D_{\text{kt}}}$, $r_{t-1}\in\{0,1\}$ \\
$\mathbf{m}_t$ & $\in \mathbb{R}^{D_{\text{history}}}$ \\
$\mathrm{MLP}_p(\cdot)$
& Linear$(D_{\text{input}} \rightarrow D_{\text{input}})$ + ReLU + Dropout + Linear$(D_{\text{input}} \rightarrow D_{\text{kt}})$ \\
\bottomrule
\end{tabularx}
\caption{Notations and tensor shapes used in the BAIM architecture.}
\label{tab:baim_notation_shapes}
\end{table}
This section describes the implementation details of the neural components in the BAIM framework, including the MLPs, recurrent modules, and routing mechanisms. A summary of all notations and tensor shapes used throughout the architecture is provided in Table~\ref{tab:baim_notation_shapes}.

\paragraph{Procedural Solution Representation.}
For each item $I_t$, the stage-aware item embeddings
$\{\mathbf{h}'_{t,p}\}_{p=0}^{3}$ are concatenated and passed through a projection
network $f_{\text{proj}}(\cdot)$ to produce a solution representation
$\mathbf{s}_t \in \mathbb{R}^{D_{\text{kt}}}$.
The projection network consists of a linear transformation
$\text{Linear}(4D_{\text{input}} \rightarrow D_{\text{kt}})$ followed by ReLU
activation and dropout.

\paragraph{Learner Interaction Context Encoder.}
At each time step $t$, the GRU updates the latent context by taking as input the
concatenation of a stage-aware solution representation derived from the previous
item and the previous response embedding $\mathbf{r}_{t-1}$.
The response embedding $\mathbf{r}_{t-1}$ is obtained from the binary response
$r_{t-1} \in \{0,1\}$ via a learnable lookup table.
The concatenated input is projected into the context space via
$\mathbf{W}_{\text{in}} \in \mathbb{R}^{D_{\text{history}} \times (2D_{\text{kt}})}$
before being fed into the GRU.

\paragraph{Routing Gate Network.}
During training, Gaussian noise is injected into the routing logits to encourage
exploration and stabilize routing behavior, following common practice in sparse
mixture-of-experts routing~\cite{shazeer2017outrageously}:
\begin{equation}
\tilde{\boldsymbol{\alpha}}_t
=
\boldsymbol{\alpha}_t + \boldsymbol{\epsilon}_t,
\quad
\boldsymbol{\epsilon}_t \sim \mathcal{N}\!\left(\mathbf{0}, (1/4)^2 \mathbf{I}\right).
\end{equation}
Routing decisions are made via a Top-1 selection based on the noisy logits
$\tilde{\boldsymbol{\alpha}}_t$, while deterministic routing using
$\boldsymbol{\alpha}_t$ is employed at inference time.

\paragraph{Stage-Specific Expert Networks.}
Each procedural reasoning stage $p \in \{0,1,2,3\}$ is associated with a lightweight
expert MLP with an identical architecture but independent parameters.
Concretely, each expert maps its corresponding stage-level embedding from
$\mathbb{R}^{D_{\text{input}}}$ to $\mathbb{R}^{D_{\text{kt}}}$ via a two-layer
feed-forward network consisting of a linear transformation
$(D_{\text{input}} \rightarrow D_{\text{input}})$, followed by ReLU activation
and dropout, and a second linear projection
$(D_{\text{input}} \rightarrow D_{\text{kt}})$.
All expert outputs are computed in parallel, and a Top-1 routing strategy is
applied such that only the selected expert output contributes to the final
representation.
A shared layer normalization is applied to the selected output, yielding the
final item representation
$\mathbf{I}_t \in \mathbb{R}^{D_{\text{kt}}}$, which is subsequently passed to the
KT backbone for response prediction.

\section{Details on Baseline Reproductions}
\label{sec:appendix_baselines}

\paragraph{Default Setting.}
In the \textbf{Default} setting, item and knowledge-component (KC) embeddings are randomly initialized and trained end-to-end together with each KT backbone. For backbones that explicitly model KC-level representations, including AKT, SimpleKT, and SparseKT, training is performed at the KC level. At inference time, item-level predictions are obtained via a late fusion strategy that aggregates KC-level outputs associated with each item. This design follows the standard usage of these backbones and ensures that all models operate under their intended training paradigms.

\paragraph{PEBG Reproduction.}
We reproduce \texttt{PEBG} following the official implementation with minimal dataset-specific adjustments.  While most architectural details and preprocessing steps strictly follow the original work, we adjust the model scale to match our experimental environment. Specifically, we set the embedding and hidden dimensions to $768$ (originally $64$ and $128$, respectively) and apply a dropout rate of $0.3$ to prevent overfitting. For items absent from the training set, we assign default attributes ($ms=0.0$, $p\_correct=0.5$) to maintain graph connectivity.

\paragraph{KCQRL Reproduction.}
We reproduce \texttt{KCQRL} using the released code and data under the original contrastive framework, yielding $768$-dimensional item embeddings. For experiments using pre-trained item embeddings from either \texttt{KCQRL} or \texttt{PEBG}, we employ item-level KT backbone variants consistent with the KCQRL architecture to enable uniform integration and evaluation.

\section{Architecture-Specific Considerations for Item Representation}
\label{sec:appendix_baim_backbones}

We integrate BAIM into existing KT backbones by modifying only the item
representation module, while preserving each model’s original
sequence modeling and prediction mechanisms.
Based on how item representations are consumed within each backbone,
we categorize the models into three groups.

\paragraph{Group A: AKT and qDKT.}
AKT and qDKT require separate embeddings for the current item and historical
item--response interactions.
Accordingly, BAIM replaces the original item and interaction embedding
modules with learner-conditioned item representations.
Specifically, BAIM produces a learner-conditioned item representation
$\mathbf{I}_t \in \mathbb{R}^{D_{\text{kt}}}$, which serves as the shared source
representation for constructing item--response embeddings.
Following the original designs of AKT and qDKT, response-aware interaction
embeddings are obtained by applying response-specific linear transformations
to $\mathbf{I}_t$, where separate projection matrices are used for correct and
incorrect responses, respectively.
In addition, a dedicated linear projection is applied to $\mathbf{I}_t$ to
construct the query embedding for the current item.
All subsequent attention, sequence modeling, and prediction components
are preserved exactly as in the original implementations.

\paragraph{Group B: QIKT.}
QIKT maintains an explicit separation between item embeddings and KC embeddings.
To respect this design, BAIM is integrated only at the item level.
The learner-conditioned item representation produced by BAIM replaces the
original item representation, while the concept embedding module and
item--concept fusion mechanism remain unchanged. Thus, BAIM does not introduce or modify any concept-level representations in QIKT.

\begin{table*}[t]
\centering
\small
\setlength{\tabcolsep}{3pt} 
\begin{tabular}{l|ccccc}
\toprule
\multirow{2}{*}{\diagbox[width=12em]{Method}{Backbone}} & \multicolumn{5}{c}{Knowledge Tracing Backbone Architecture} \\
\cmidrule(l){2-6}
 & AKT & qDKT & QIKT & simpleKT & sparseKT \\
\midrule

\multicolumn{6}{l}{\textbf{XES3G5M}} \\
\midrule
InternVL-3.5-8B
    & {82.95} {\scriptsize $\pm$ 0.04}
    & {82.39} {\scriptsize $\pm$ 0.02}
    & {82.15} {\scriptsize $\pm$ 0.02}
    & {82.81} {\scriptsize $\pm$ 0.03}
    & {83.17} {\scriptsize $\pm$ 0.09}\\
Qwen3-VL-8B-Thinking 
    & {82.95} {\scriptsize $\pm$ 0.05}
    & {82.39} {\scriptsize $\pm$ 0.02}
    & {82.14} {\scriptsize $\pm$ 0.03}
    & \textbf{82.84} {\scriptsize $\pm$ 0.01}
    & \textbf{83.22} {\scriptsize $\pm$ 0.08}\\
Qwen3-VL-32B-Thinking 
    & \textbf{83.00} {\scriptsize $\pm$ 0.04}
    & \textbf{82.43} {\scriptsize $\pm$ 0.02}
    & \textbf{82.17} {\scriptsize $\pm$ 0.05}
    & \textbf{82.84} {\scriptsize $\pm$ 0.01}
    & {83.21} {\scriptsize $\pm$ 0.10}\\
\midrule

\multicolumn{6}{l}{\textbf{NIPS34}} \\
\midrule
InternVL-3.5-8B
    & {80.12} {\scriptsize $\pm$ 0.06}
    & {80.11} {\scriptsize $\pm$ 0.01}
    & {80.16} {\scriptsize $\pm$ 0.05}
    & {79.99} {\scriptsize $\pm$ 0.03}
    & {80.34} {\scriptsize $\pm$ 0.09}\\
Qwen3-VL-8B-Thinking 
    & {80.14} {\scriptsize $\pm$ 0.03}
    & \textbf{80.15} {\scriptsize $\pm$ 0.03}
    & \textbf{80.18} {\scriptsize $\pm$ 0.01}
    & \textbf{80.02} {\scriptsize $\pm$ 0.02}
    & \textbf{80.45} {\scriptsize $\pm$ 0.07}\\
Qwen3-VL-32B-Thinking 
    & \textbf{80.16} {\scriptsize $\pm$ 0.04}
    & {80.13} {\scriptsize $\pm$ 0.03}
    & \textbf{80.18} {\scriptsize $\pm$ 0.04}
    & \textbf{80.02} {\scriptsize $\pm$ 0.03}
    & {80.36} {\scriptsize $\pm$ 0.12}\\
\bottomrule
\end{tabular}
\caption{
Performance comparison (mean $\pm$ std) of BAIM-equipped KT backbones using different solver RLMs.
The best-performing solver in terms of mean AUC is highlighted in \textbf{bold}.
}
\label{tab:main_result_family}
\end{table*}

\paragraph{Group C: simpleKT and sparseKT.}
simpleKT and sparseKT employ single-stream Transformer architectures in which
item embeddings are directly used as query representations.
For these models, BAIM replaces the original static item embeddings with
learner-conditioned item representations projected to $D_{\text{kt}}$.
Following the original designs of simpleKT and sparseKT, response information
is incorporated by directly combining the learner-conditioned item
representation with the response embedding, rather than through
response-specific linear projections.
In sparseKT, the original sparse attention mechanism and its hyperparameters
are fully preserved; BAIM only affects the input embedding supplied to the
Transformer. This ensures that the sparsification behavior remains identical to the
original implementation.

\section{Data Preprocessing Details}
\label{sec:appendix_preprocessing}

In this section, we provide detailed descriptions of the preprocessing pipelines for the two benchmark datasets used in our study to ensure reproducibility and transparency in our procedural solution extraction process.

\paragraph{XES3G5M Metadata Translation.}
The XES3G5M dataset was collected from a Chinese online learning platform, where all question texts and analytical analyses are provided in Chinese. Although the underlying solver RLM supports Chinese, we translate the metadata into English to standardize the working language of our analysis pipeline and to enable more reliable human inspection, error analysis, and qualitative evaluation. Translation is performed using the GPT-5-nano model~\cite{singh2025openai}, with particular care taken to preserve mathematical notation and the logical structure of the original analyses.

During the translation process, we identified indexing issues, particularly cases where image file names were included in the option fields. These entries are treated as annotation noise originating from the source dataset. The manually corrected metadata is used in all our experiments, and the finalized version is fully available in our public repository.

\paragraph{NIPS34 Image-based Metadata Generation}
Since the NIPS34 provides only question images without structured text or analysis, we generate metadata using a \textbf{Gemini-2.5-Pro}~\cite{comanici2025gemini}. 
Given an input image, the model jointly performs visual understanding
and text generation to extract the question content, multiple-choice options, a concise analytical explanation, and the correct answer. The full prompt used for metadata generation is shown in Figure~\ref{fig:metadata_prompt_single}.

\begin{table}[ht]
    \centering
    \label{tab:token_stats_comparison}
    \resizebox{\columnwidth}{!}{ 
    \begin{tabular}{lcccc}
        \toprule
        \textbf{Stage} & \textbf{Mean} & \textbf{Std Dev} & \textbf{Min} & \textbf{Max} \\
        \midrule
        \multicolumn{5}{l}{\textit{\textbf{Qwen3-VL-32B-Thinking}}} \\
        \midrule
        \textbf{Thinking Process} & \textbf{1,191.78} & \textbf{1,107.05} & \textbf{206} & \textbf{11,704} \\
        Stage 1: Understand & 50.05 & 15.29 & 16 & 149 \\
        Stage 2: Plan & 43.39 & 14.23 & 9 & 140 \\
        Stage 3: Carry Out & 76.80 & 40.49 & 7 & 375 \\
        Stage 4: Look Back & 36.52 & 25.20 & 9 & 1,773 \\
        \textit{Total Sequence} & 1,423.48 & 1,128.02 & 340 & 11,999 \\
        \midrule
        \multicolumn{5}{l}{\textit{\textbf{Qwen3-VL-8B-Thinking}}} \\
        \midrule
        \textbf{Thinking Process} & \textbf{2,064.72} & \textbf{2,245.83} & \textbf{214} & \textbf{17,655} \\
        Stage 1: Understand & 56.28 & 17.87 & 16 & 198 \\
        Stage 2: Plan & 50.09 & 22.56 & 12 & 348 \\
        Stage 3: Carry Out & 96.15 & 63.31 & 9 & 892 \\
        Stage 4: Look Back & 50.66 & 24.12 & 10 & 275 \\
        \textit{Total Sequence} & 2,344.07 & 2,286.52 & 354 & 17,941 \\
        \midrule
        \multicolumn{5}{l}{\textit{\textbf{InternVL-3.5-8B}}} \\
        \midrule
        \textbf{Thinking Process} & \textbf{1,232.83} & \textbf{1,569.69} & \textbf{46} & \textbf{11,628} \\
        Stage 1: Understand & 39.51 & 13.81 & 9 & 161 \\
        Stage 2: Plan & 34.54 & 15.90 & 8 & 273 \\
        Stage 3: Carry Out & 83.24 & 55.55 & 10 & 1,786 \\
        Stage 4: Look Back & 30.56 & 17.13 & 4 & 689 \\
        \textit{Total Sequence} & 1,420.69 & 1,589.12 & 125 & 11,855 \\
        \bottomrule
    \end{tabular}
    }
    \caption{
    Token usage statistics for solver RLM generation across model variants on the XES3G5M.
    }
    \label{tab:gen_token_stats}
\end{table}

The generated analytical explanations serve as reference material
for downstream reasoning extraction, enabling the solver module
to derive stage-wise problem-solving trajectories.
This preprocessing step allows BAIM to be applied to image-based educational datasets
that do not natively provide textual problem descriptions or procedural solution traces.

\paragraph{Interaction Filtering and Splitting.}
Following the standard protocol of the \texttt{pyKT} library, all datasets were divided into training, validation, and test sets using the default splitting ratios to facilitate fair comparison with existing baselines.

\section{Hardware Usage}
For our experiments, we used a single NVIDIA GeForce RTX 3090 GPU for training the KT models, and two NVIDIA L40S GPUs for RLM inference.

\section{RLM-Family Analysis}
\label{sec:appendix_model_family}

Table~\ref{tab:main_result_family} shows that BAIM consistently improves
performance across all evaluated KT backbones on XES3G5M and NIPS34.
Across different solver families, the overall performance remains comparable,
indicating that BAIM is robust to the choice of solver model.
In particular, the performance gap between Qwen3-VL-8B-Thinking and
Qwen3-VL-32B-Thinking is marginal across most backbones, suggesting that the
procedural information extracted by the solver does not strongly depend on
model scale once a sufficient capacity threshold is reached.

In practice, we adopt Qwen3-VL-32B-Thinking in the main experiments because Qwen3-VL-8B-Thinking more frequently exhibits overthinking, producing unnecessarily long reasoning traces. While downstream performance remains comparable, this behavior increases preprocessing cost and reduces generation efficiency. Detailed token usage statistics are reported in Table~\ref{tab:gen_token_stats}.

\begin{table}[t]
\centering
\small
\setlength{\tabcolsep}{4pt}
\renewcommand{\arraystretch}{1.05}
\resizebox{\columnwidth}{!}{%
\begin{tabular}{c c c c c c}
\toprule
$\lambda$ & AUC & Stage 0 & Stage 1 & Stage 2 & Stage 3 \\
\midrule
0    & $83.18 \pm 0.06$ & 0.48 & 0.44 & 0.08 & 0.01 \\
0.01 & $\mathbf{83.21 \pm 0.10}$ & 0.24 & 0.29 & 0.23 & 0.24 \\
0.1  & $83.16 \pm 0.19$ & 0.28 & 0.23 & 0.31 & 0.18 \\
\bottomrule
\end{tabular}%
}
\caption{Effect of the load-balancing coefficient $\lambda$ on XES3G5M with sparseKT. We report test AUC and aggregated Top-1 routing proportions over the four procedural stages.}
\label{tab:lambda_ablation}
\end{table}
\section{Effect of the Load-Balancing Loss}
\label{sec:appendix_lambda}

We analyze the effect of the load-balancing regularizer by varying its coefficient
$\lambda \in \{0, 0.01, 0.1\}$ on XES3G5M with the sparseKT backbone.
As shown in Table~\ref{tab:lambda_ablation}, the overall AUC is similar across settings, with $\lambda=0.01$ achieving the best performance.
However, the routing behavior differs substantially.
Without load balancing ($\lambda=0$), Top-1 routing collapses to a small subset of stages, with about $92\%$ of samples assigned to Stage 0 or Stage 1.
In contrast, $\lambda=0.01$ yields a much more balanced routing distribution across all four stages, while preserving the best AUC.
A larger value, $\lambda=0.1$, also mitigates collapse but gives slightly lower performance. These results suggest that an appropriate choice of $\lambda$ can effectively prevent stage collapse while also providing a modest improvement in AUC.

\section{Solver RLM Inference}
\label{sec:solver_rlm_inference}
\paragraph{Decoding Hyperparameters.} Because we observed overthinking behavior in Qwen3-VL-8B-Thinking, we set \texttt{max\_tokens}=18000 for Qwen3-VL-8B-Thinking, and \texttt{max\_tokens}=12000 for Qwen3-VL-32B-Thinking and InternVL-3.5-8B~\cite{wang2025internvl3_5}. Other hyperparameters are fixed to temperature = \texttt{0.7}, top-$p$ = \texttt{0.9}, and repetition penalty = \texttt{1.1}.

\paragraph{Prompting Setup.}
We use a fixed prompt to elicit Polya-style four-stage reasoning in JSON format. The full prompt is shown in Figure~\ref{fig:solver_prompt}.
\label{sec:solver_prompt}

\paragraph{Robustness to RLM Reasoning Errors.}
To evaluate the reliability of the solver RLM, we manually inspected $1,000$ randomly sampled outputs of Qwen3-VL-32B-Thinking on the XES3G5M. We identified that only $1.3\%$ of the cases exhibited logical inconsistencies or incorrect final answers, typically occurring when the source metadata was highly ambiguous or the reference analysis was overly concise.

\clearpage

\begin{figure*}[p]
\centering
\begin{promptbox}

\textbf{SYSTEM PROMPT}
\vspace{0.1cm}

You are a math education expert. Your task is to read the provided question image and extract its content into a structured JSON representation suitable for downstream reasoning.

\vspace{0.2cm}
\textbf{Output Format Rules}
\begin{itemize}[leftmargin=*]
    \item Output \textbf{one valid JSON object only} (no additional text).
    \item The JSON must contain exactly the following fields:
    \begin{itemize}
        \item \texttt{"question"}: Concise textual description.
        \item \texttt{"options"}: Dict with keys \texttt{"A", "B", "C", "D"}.
        \item \texttt{"analysis"}: Brief explanation (3--4 sentences).
        \item \texttt{"answer"}: Correct option key.
    \end{itemize}
\end{itemize}

\vspace{0.2cm}
\textbf{Content Constraints}
\begin{itemize}[leftmargin=*]
    \item Use \LaTeX\ for all mathematical expressions.
    \item Reference figures as \texttt{question\_\{id\}-image\_0}.
    \item Ensure JSON is syntactically valid.
\end{itemize}

\hrule
\vspace{0.2cm}

\textbf{INSTRUCTION}
\vspace{0.1cm}

Process the given question image and generate structured metadata that faithfully captures the problem statement, solution logic, and correct answer.

\vspace{0.2cm}
\textbf{IMPORTANT}
\begin{itemize}[leftmargin=*]
    \item Do not include extraneous commentary.
    \item Follow the specified JSON schema strictly.
\end{itemize}

\vspace{0.2cm}
Image: \texttt{\{question\_image\}}

\end{promptbox}
\caption{Prompt used to generate structured metadata from question images using Gemini-2.5-Pro.}
\label{fig:metadata_prompt_single}
\end{figure*}

\clearpage
\begin{figure*}[p]
\centering
\begin{promptbox}

% --- Shared System Prompt ---
\textbf{SHARED SYSTEM PROMPT}

You are a math student solving problems using \textbf{Polya's 4-Stage Process}.
You must explain your reasoning \textbf{in English only}.

\textbf{Output Format Rules}
\begin{itemize}[leftmargin=*]
    \item Immediately after \texttt{</think>}, output \textbf{one JSON object only} (no extra text).
    \item The JSON must have exactly these 4 keys:
    \begin{enumerate}
        \item \texttt{"understand"}: Identify knowns, unknowns, and conditions.
        \item \texttt{"plan"}: State the strategy, formulas, or theorems to be used.
        \item \texttt{"carry\_out"}: Execute the plan step-by-step with calculations.
        \item \texttt{"look\_back"}: State the final answer and verify if it makes sense.
    \end{enumerate}
\end{itemize}
\textbf{Content Constraints}
\begin{itemize}[leftmargin=*]
    \item Keep sentences concise and logical.
    \item Ensure the JSON is valid and parseable.
\end{itemize}

\textbf{Example Output}
\begin{tcolorbox}[colback=white, colframe=gray!50, sharp corners]
\small
\begin{verbatim}
{
  "understand": "Given a circle with radius 5. Need to find the area.",
  "plan": "Use the area formula A = pi * r^2. Substitute r=5.",
  "carry_out": "A = pi * 5^2 = 25 * pi. Approx 78.5.",
  "look_back": "The result 25pi is positive and reasonable. Final Answer: 25pi"
}
\end{verbatim}
\end{tcolorbox}
\hrule
\vspace{0.3cm}

% --- Instruction C ---
\textbf{INSTRUCTION}

Simulate a \textbf{high-performing student}.
Solve the following problem by strictly following Polya's 4 stages correctly.

\vspace{0.2cm}

\textbf{IMPORTANT}: You are provided with a reference \textbf{Analysis}.
\begin{itemize}[leftmargin=*]
    \item Do not just copy the Analysis.
    \item \textbf{Reconstruct} the reasoning path based on the Analysis to ensure your steps and calculations are 100\% accurate.
\end{itemize}

\begin{itemize}[leftmargin=*]
    \item \textbf{understand}: Extract key information from the Question.
    \item \textbf{plan}: Formulate a strategy consistent with the provided Analysis.
    \item \textbf{carry\_out}: Perform precise calculations step-by-step as outlined in the Analysis.
    \item \textbf{look\_back}: Verify the result matches the Analysis's conclusion.
\end{itemize}

\vspace{0.1cm}
Question:
\texttt{\{question\}}

\vspace{0.1cm}
Analysis:
\texttt{\{analysis\}}

\end{promptbox}
\caption{System prompt used to elicit Polya-style four-stage reasoning trajectories from the solver RLM.}
\label{fig:solver_prompt}
\end{figure*}

\end{document}